# Multi-Range Attentive Bicomponent Graph Convolutional Network for Traffic Forecasting

**Weiqi Chen[1], Ling Chen[1, *], Yu Xie[2], Wei Cao[2], Yusong Gao[2], Xiaojie Feng[2]**

[1] College of Computer Science and Technology, Zhejiang University, Hangzhou 310027, China
[2] Alibaba Group, Hangzhou 311121, China
vc12301@gmail.com, lingchen@cs.zju.edu.cn,
{qianqing.xy, mingsong.cw, jianchuan.gys, xiaojie.fxj}@alibaba-inc.com

**Abstract**

Traffic forecasting is of great importance to transportation management and public safety, and very challenging due to the complicated spatial-temporal dependency and essential uncertainty brought about by the road network and traffic conditions. Latest studies mainly focus on modeling the spatial dependency by utilizing graph convolutional networks (GCNs) throughout a fixed weighted graph. However, edges, i.e., the correlations between pair-wise nodes, are much more complicated and interact with each other. In this paper, we propose the *Multi-Range Attentive Bicomponent GCN* (MRA-BGCN), a novel deep learning model for traffic forecasting. We first build the node-wise graph according to the road network distance and the edge-wise graph according to various edge interaction patterns. Then, we implement the interactions of both nodes and edges using bicomponent graph convolution. The multi-range attention mechanism is introduced to aggregate information in different neighborhood ranges and automatically learn the importance of different ranges. Extensive experiments on two real-world road network traffic datasets, METR-LA and PEMS-BAY, show that our MRA-BGCN achieves the state-of-the-art results.

# Introduction

Traffic forecasting is one of the most challenging tasks in Intelligent Transportation System (ITS) (Jabbarpour et al. 2018) and of great importance to transportation management and public safety. The task of traffic forecasting is to forecast the future traffic of a road network given the historical traffic data.

This task is very challenging mainly due to the complicated spatial-temporal dependency and essential uncertainty brought about by the road network and traffic conditions. On the one hand, the irregular underlying road network results in complicated correlations among traffic data. On the other hand, due to various unpredictable traffic conditions, traffic data is inherently uncertain.

* Corresponding Author

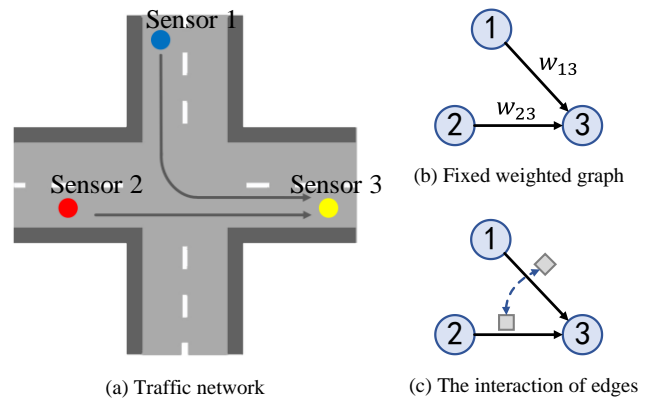

Figure 1: The Interaction of Edges in a Node-Wise Graph

Early traffic forecasting approaches (Nikovski et al., 2005; Chien et al., 2003; Lippi et al., 2013) mainly employ shallow machine learning for a single observation node or few nodes, which are limited by the capability of capturing the nonlinearity in traffic data and neglect or barely leverage the spatial dependency. Recent advances in deep learning make it possible to model the complicated spatial-temporal dependency in traffic forecasting. Some attempts (Ma et al., 2017; Zhao et al., 2017; Zhang et al., 2018) applied Convolutional Neural Networks (CNNs) and Recurrent Neural Networks (RNNs) for traffic forecasting. However, CNNs restrict the model to process grid structures (e.g., images and videos), and the non-Euclidean correlations dominated by irregular road networks are not considered. To tackle this problem, Graph Convolutional Networks (GCNs), which are efficient in handling non-Euclidean correlations, are integrated with RNNs (Li et al., 2018) or CNNs (Yu et al., 2018) to embed the prior knowledge of the road network and capture the correlations between pair-wise nodes. Despite promising results of in-troducing GCNs, we argue that there are still two important aspects neglected in these approaches.

First, these approaches mainly focus on modeling the spatial dependency by utilizing GCNs throughout a fixed weighted graph. However, edges, i.e., the correlations between pair-wise nodes, are much more complicated and interact with each other. Figure 1 illustrates an example. As shown in Figure 1(a), sensors 1 and 3, as well as sensors 2 and 3, are correlated by road links. Obviously, these correlations change with the current traffic condition and interact with each other. As shown in Figure 1(b), existing approaches build a weighted graph according to the road network distance and use a GCN to implement the interaction of nodes, while the correlations between pair-wise nodes are represented by fixed scalars in the adjacency matrix, which neglects the complexity and interaction of edges.

Second, these approaches usually use the information aggregated in a given neighborhood range (i.e., neighbors within $k$-hops), ignoring multiple range information. However, information in different ranges reveals distinct traffic properties. A small neighborhood range indicates the local dependency, and a large range tends to uncover an overall traffic pattern in a relatively large region. Furthermore, information in different ranges does not contribute equally in all cases. For example, due to a traffic accident, a node is predominantly influenced by its nearest neighbors, on which a model should pay more attention rather than considering all neighbors within $k$-hops equally.

To address the aforementioned problems, we propose a deep learning model called *Multi-Range Attentive Bicomponent GCN* (MRA-BGCN), which not only considers node correlations, but also regards edges as entities that interact with each other, as shown in Figure 1(c), and leverages multiple range information. The main contributions of our work are as follows:

- We propose MRA-BGCN, which introduces the bicomponent graph convolution to explicitly model the correlations of both nodes and edges. The node-wise graph is built according to the road network distance, and the edge-wise graph is built by considering two types of edge interaction patterns, stream connectivity and competitive relationship.

- We propose the multi-range attention mechanism for the bicomponent graph convolution, which can aggregate information in different neighborhood ranges and learn the importance of different ranges.

- We conduct extensive experiments on two real-world traffic datasets, METR-LA and PEMS-BAY, and the proposed model achieves the state-of-the-art results.

## Related Works

Early traffic forecasting approaches, e.g., Linear Regression based approach (Nikovski et al., 2005), Kalman Filtering based approach (Chien et al., 2003), and Auto-Regressive Integrated Moving Average (ARIMA) based approach (Lippi et al., 2013), mainly employ shallow machine learning for a single observation node or few nodes, which are limited by the capability of capturing the nonlinearity in traffic data and neglect or barely leverage the spatial dependency.

Recent advances in deep learning make it possible to model the complicated spatial-temporal dependency in traffic forecasting. Some attempts (Ma et al., 2017; Zhao et al., 2017; Zhang et al. 2018) applied Convolutional Neural Networks (CNNs) and Recurrent Neural Networks (RNNs) for traffic forecasting. In these studies, CNNs, which are restricted to processing regular grid structures (e.g., images and videos), are introduced to capture the spatial dependency, while the non-Euclidean correlations dominated by irregular road networks are not considered.

To tackle this problem, researchers have applied graph convolution to model the non-Euclidean correlations for traffic forecasting. Li et al. (2018) proposed Diffusion Convolutional Recurrent Neural Network (DCRNN), which replaces the fully-connected layers in Gated Recurrent Units (GRU) (Chung et al., 2014) by the diffusion convolution operator. The diffusion convolution performs graph convolution on the given graph and its inverse to consider both inflow and outflow relationships. Yu et al. (2018) proposed Spatial-Temporal GCN (ST-GCN), which combines a graph convolution with a 1D convolution. In ST-GCN, the graph convolution captures the spatial dependency, and the 1D convolution is employed on time axis to capture the temporal dependency, which is much more computationally efficient than RNNs.

The above-mentioned GCN-based approaches encode the road network distance into a fixed weighted graph representing the spatial dependency. To further modeling the complicated correlations in traffic forecasting, Wu et al. (2019a) proposed to capture the hidden spatial dependency that is unseen in the given graph with a self-adaptive adjacency matrix. This self-adaptive adjacency matrix is achieved by computing the similarity of node embeddings. However, the hidden spatial dependency is learnt in a data-driven manner, which lacks the guidance of the domain knowledge and may suffer from the overfitting problem. In addition, existing traffic forecasting approaches are ineffective to model the interaction of edges and leverage multiple range information.

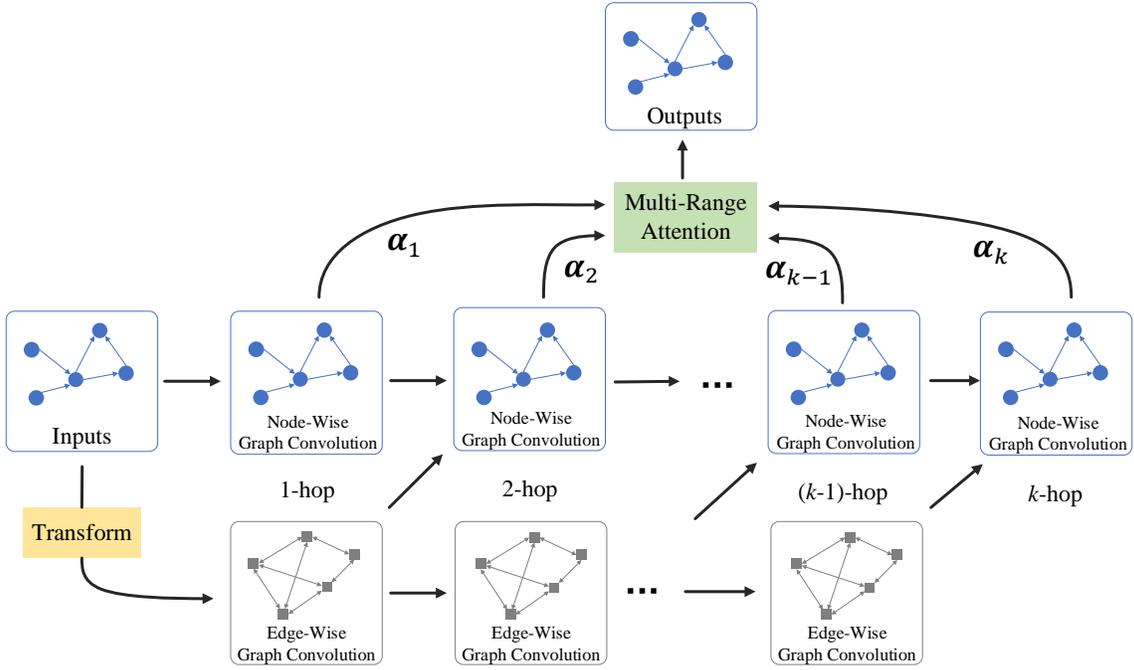

Figure2: The Architecture of MRA-BGCN

## Preliminaries

### Problem Definition

Given the historical traffic data from $N$ correlated traffic sensors located on a road network, the task of traffic forecasting is to forecast the future traffic of the road network. Following previous studies, we define the $N$ correlated traffic sensors as a weighted directed graph $G = (V, E, A)$, where $V$ is a set of $|V| = N$ nodes, $E$ is a set of edges, and $A \in \mathbb{R}^{N \times N}$ is a weighted adjacency matrix representing the nodes' proximities, e.g., the road network distance between any pair of nodes. The traffic data observed on $G$ at time $t$ are denoted as a graph signal $X^{(t)} \in \mathbb{R}^{N \times P}$, where $P$ is the feature dimension of each node. The traffic forecasting problem aims to learn a function $f$ that is able to forecast $T$ future graph signals given $T'$ historical graph signals and the graph $G$:

$$[X^{(t-T'+1):t}, G] \xrightarrow{f} [X^{(t+1):(t+T)}],$$

where $X^{(t-T'+1):t} \in \mathbb{R}^{N \times P \times T'}$ and $X^{(t+1):(t+T)} \in \mathbb{R}^{N \times P \times T}$.

### Graph Convolution

GCNs are building blocks for learning data with non-Euclidean structures, i.e., graphs (Wu et al., 2019b). They are widely applied in node classification (Kipf and Welling, 2017), graph classification (Ying et al., 2018), link prediction (Zhang and Chen, 2018), etc. GCN approaches fall into two categories, spectral-based and spatial-based. Spectral-based approaches conduct graph Fourier transformation and apply convolutional filters on the spectral domain (Defferrard et al., 2016; Kipf and Welling, 2017). Spatial-based approaches aggregate the representations of a node and its neighbors to get a new representation for the node (Atwood and Towsley, 2016; Gilmer et al., 2017; Hamilton et al., 2017; Velickovic et al. 2017).

We briefly describe the graph convolution operator applied in our model. A graph convolution is defined over a graph $G = (V, E, A)$:

$$\theta_{\star_G} X = \rho(\widetilde{D}^{-1}\widetilde{A}X\theta), \qquad (1)$$

where $X \in \mathbb{R}^{N \times P}$ is the input signal, $\theta \in \mathbb{R}^{P \times F}$ is the learnable parameter matrix, $\widetilde{A} = A + I_N$ is the adjacency matrix with self-connection, $\widetilde{D}$ is the diagonal degree matrix of $\widetilde{A}$, $\widetilde{D}^{-1}\widetilde{A}$ represents the normalized adjacency matrix, and $\rho$ is a nonlinear activation function. A graph convolution can aggregate information of 1-hop neighbors. By stacking multiple graph convolution layers, we can expand the receptive neighborhood range.

## Methodology

### Model Overview

Figure 2 demonstrates the architecture of MRA-BGCN, which consists of two parts: (1) the bicomponent graph convolution module; and (2) the multi-range attention layer. The bicomponent graph convolution module contains sev-eral node-wise graph convolution layers and edge-wise graph convolution layers, which can explicitly model the interactions of both nodes and edges. The multi-range attention layer aggregates information in different neighborhood ranges and learns the importance of different ranges. In addition, we combine MRA-BGCN with RNN to model the temporal dependency for traffic forecasting. The detailed implementation is described in the following subsections.

## Bicomponent Graph Convolution

Graph convolution is an efficient operation to model the interaction of nodes given the graph structure. However, in traffic forecasting, edges, i.e., the correlations between pair-wise nodes, are much more complicated and interact with each other. Thus, we propose the bicomponent graph convolution, which can explicitly model the interactions of both nodes and edges.

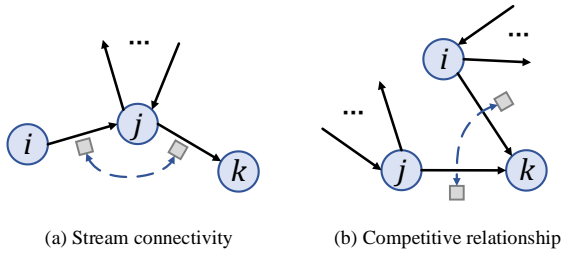

(a) Stream connectivity  (b) Competitive relationship

Figure 3: Edge Interaction Patterns

Chen et al. (2019) proposed to introduce line graph of edge adjacencies to model edge correlations. Let $G = (V, E, \mathbf{A})$ denotes the node-wise directed graph. $G_L = (V_L, E_L, \mathbf{A}_L)$ is the corresponding line graph, then the nodes $V_L$ of $G_L$ are the ordered edges in $E$, i.e., $V_L = \{(i \to j); (i, j) \in E\}$ and $|V_L| = |E|$. $\mathbf{A}_L$ is an unweighted adjacency matrix that encodes the edge adjacencies in the node-wise graph, which is defined as: $\mathbf{A}_{L,(i \to j),(j \to k)} = 1$ and 0 otherwise.

Despite the capability of considering edge adjacencies, the line graph is an unweighted graph and only considers two edges are correlated if one's target node shares with the other one's source node. However, it is ineffective to characterize various edge interaction patterns that are common in traffic forecasting. As shown in Figure 3, we define two types of edge interaction patterns to construct the edge-wise graph $G_e = (V_e, E_e, \mathbf{A}_e)$. Note that, each node of $V_e$ represents an edge of $E$.

**Stream connectivity**: In a traffic network, a road link is possibly influenced by its upstream and downstream road links. As shown in Figure 3(a), $(i \to j)$ is an upstream edge of $(j \to k)$, and thus they are correlated. Intuitively, if the joint node $j$ has a large number of neighbors (i.e., the degree of $j$ is large), the correlation between $(i \to j)$ and $(j \to k)$ is weak, as it is susceptible to other neighbors. We compute the edge weights for the stream connectivity in $\mathbf{A}_e$ using Gaussian kernel:

$$\mathbf{A}_{e,(i \to j),(j \to k)} = \mathbf{A}_{e,(j \to k),(i \to j)} = \exp\left(-\frac{(\deg^-(j) + \deg^+(j) - 2)^2}{\sigma^2}\right), \quad (2)$$

where $\deg^-(j)$ and $\deg^+(j)$ denote the indegree and outdegree of node $j$ in the node-wise graph, respectively, and $\sigma$ is the standard deviation of node degrees.

**Competitive relationship**: Road links sharing a same source node probably contend for traffic resources and incur competitive relationship. As shown in Figure 3(b), two edges, $(i \to k)$ and $(j \to k)$, sharing the target node $k$ are correlated due to competitive relationship. Analogous to stream connectivity, the intensity of competitive relationship is related to the outdegrees of the source nodes. For example, if the source node of an edge has multiple outcoming edges, this edge is robust for the competition of traffic resources. Thus, we compute the edge weights for the competitive relationship in $\mathbf{A}_e$ as:

$$\mathbf{A}_{e,(i \to k),(j \to k)} = \mathbf{A}_{e,(j \to k),(i \to k)} = \exp\left(-\frac{(\deg^+(i) + \deg^+(j) - 2)^2}{\sigma^2}\right). \quad (3)$$

With the constructed edge-wise graph $G_e$, as shown in Figure 2, the bicomponent graph convolution can explicitly model the interactions of both nodes and edges. The $k$-hop bicomponent graph convolution is formulated as:

$$\begin{aligned} \mathbf{X}^{(l+1)} &= \boldsymbol{\theta}_{n}^{(l)} {}_{\star G} [\mathbf{X}^{(l)}, \mathbf{M}\mathbf{Z}^{(l)}] \ \text{ for } l = 1, \cdots, k-1, \\ \mathbf{Z}^{(l+1)} &= \boldsymbol{\theta}_{e}^{(l)} {}_{\star G} \mathbf{Z}^{(l)} \ \text{ for } l = 0, \cdots, k-1, \\ \mathbf{X}^{(1)} &= \boldsymbol{\theta}_{n}^{(0)} {}_{\star G} \mathbf{X}^{(0)}, \\ \mathbf{Z}^{(0)} &= \mathbf{M}^{\mathrm{T}} \mathbf{X}^{(0)} \mathbf{W}_{\mathrm{b}}, \end{aligned} \quad (4)$$

where $\boldsymbol{\theta}_{\star G}$ is the graph convolution operation with parameter $\boldsymbol{\theta}$, $[\cdot, \cdot]$ is the concatenation operation, $\mathbf{X}^{(l-1)}$ is the input to layer $l$ of the node-wise graph convolution, $\mathbf{Z}^{(l-1)}$ is the input to layer $l$ of the edge-wise graph convolution, $\mathbf{M} \in \mathbb{R}^{|V| \times |E|}$ is the incidence matrix that encodes the connections between nodes and edges, defined as: $\mathbf{M}_{i,(i \to j)} = \mathbf{M}_{j,(i \to j)} = 1$ and 0 otherwise. $\mathbf{M}\mathbf{Z}^{(\cdot)}$ aggregates edge representations connected with each single node, and $\mathbf{M}^{\mathrm{T}} \mathbf{X}^{(\cdot)}$ aggregates node representations connected with each single edge. $\mathbf{W}_{\mathrm{b}}$ is a learnable projection matrix that transforms the original node input $\mathbf{X}^{(0)}$ to the original edge input $\mathbf{Z}^{(0)}$.

## Multi-Range Attention

We propose the multi-range attention mechanism for the bicomponent graph convolution to automatically learn the importance of different neighborhood ranges, which is capable of aggregating information in different neighborhood ranges rather than the given neighborhood range (i.e., neighbors within $k$-hops) only.

The bicomponent graph convolution module obtains node representations in different neighborhood ranges, $\mathcal{X} = \{X^{(1)}, X^{(2)}, \cdots, X^{(k)}\}$, $X^{(l)} \in \mathbb{R}^{|V| \times F}$, where $k$ is the maximum hop (i.e., the number of layers in the bicomponent graph convolution module), and $F$ is the representation dimension of each node. $X_i^{(l)} \in \mathbb{R}^F$ denotes node $i$'s representation in layer $l$. The multi-range attention layer aims to capture an integrated representation from multiple neighborhood ranges. To this end, first, a shared linear transformation, parameterized by $W_a \in \mathbb{R}^{F \times F'}$, is applied to every node in each layer. Then, the attention coefficients of each layer are measured by calculating the similarity of $W_a X_i^{(l)}$ and $u$, where $u \in \mathbb{R}^{F'}$ is the neighborhood range context embedding, which is initialized as a random vector and jointly learnt during the training process. Finally, the SoftMax function is applied to normalize the coefficients. The multi-range attention mechanism is formulated as:

$$e_i^{(l)} = (W_a X_i^{(l)})^T u,$$
$$a_i^{(l)} = \text{SoftMax}_l(e_i^{(l)}) = \frac{\exp(e_i^{(l)})}{\sum_{l=1}^k \exp(e_i^{(l)})}. \quad (5)$$

Once the normalized attention coefficients are obtained, we compute a linear combination of representations in each layer for every node as:

$$h_i = \sum_{l=1}^k a_i^{(l)} X_i^{(l)}. \quad (6)$$

### Bicomponent Graph Convolutional RNN

RNNs have shown impressive capability of modeling the temporal dependency. Following Seo et al. (2018) and Li et al. (2018), we combine the proposed MRA-BGCN with GRU (Chung et al., 2014) by replacing the fully-connected layers in GRU with MRA-BGCN. We refer this RNN structure as Bicomponent Graph Convolutional GRU (BGCGRU). To simplify notations, we denote $\mathcal{G}(X; \Theta)$ as applying MRA-BGCN to the input $X$, and $\Theta$ is the total trainable parameters. Then, BGCGRU is formulated as:

$$z^{(t)} = \sigma(\mathcal{G}([X^{(t)}, H^{(t-1)}]; \Theta_z)),$$
$$r^{(t)} = \sigma(\mathcal{G}([X^{(t)}, H^{(t-1)}]; \Theta_r)),$$
$$C^{(t)} = \tanh(\mathcal{G}([X^{(t)}, (r^{(t)} \odot H^{(t-1)})]; \Theta_c)), \quad (7)$$
$$H^{(t)} = z^{(t)} \odot H^{(t-1)} + (1 - z^{(t)}) \odot C^{(t)},$$

where $X^{(t)}$ and $H^{(t)}$ denote the input and output at time step $t$, $z^{(t)}$ and $r^{(t)}$ denote the update gate and reset gate at time step $t$, $\sigma$ is the Sigmoid function, and $\odot$ is the Hadamard product. As shown in Figure 4, we stack several BGCGRU layers and employ the Sequence to Sequence architecture (Sutskever et al., 2014) for multiple step ahead traffic forecasting.

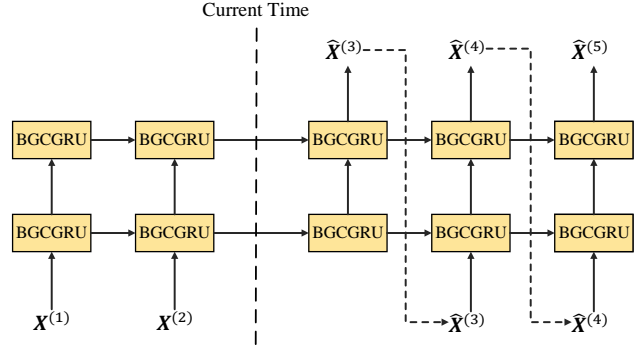

Figure 4: The Sequence to Sequence Architecture

Table 1: The Statistics of METR-LA and PEMS-BAY

| Dataset  | #Nodes | #Edges | #Time Steps |
|----------|--------|--------|-------------|
| METR-LA  | 207    | 1515   | 34272       |
| PEMS-BAY | 325    | 2369   | 52116       |

## Experiments

### Datasets

We evaluate MRA-BGCN on two public traffic network datasets, METR-LA and PEMS-BAY (Li et al., 2018). METR-LA records four months of statistics on traffic speed, ranging from Mar 1st 2012 to Jan 30th 2012, including 207 sensors on the highways of Los Angeles County. PEMS-BAY contains six months of statistics on traffic speed, ranging from Jan 1st 2017 to May 31th 2017, including 325 sensors in the Bay area. We adopt the same data pre-processing procedures as Li et al. (2018). The observations of the sensors are aggregated into 5-minute windows. The adjacency matrix of the node-wise graph is constructed by road network distance with thresholded Gaussian kernel (Shuman et al., 2013). Z-score normalization is applied to the inputs. Both the datasets are split in chronological order with 70% for training, 10% for validation, and 20% for testing. Detailed statistics of the datasets are shown in Table 1.

### Baselines

We compare MRA-BGCN with the following models:

- HA: Historical Average, which models the traffic flow as a seasonal process, and uses the average previous seasons as the prediction. The period is set to 1 week and the prediction is based on the traffic data at the same time in previous weeks.

Table2: The Performance Comparison of Multiple Step Ahead Traffic Forecasting

| Dataset | models | 15 min | | | 30 min | | | 1 hour | | |
|---|---|---|---|---|---|---|---|---|---|---|
| | | MAE | RMSE | MAPE | MAE | RMSE | MAPE | MAE | RMSE | MAPE |
| METR-LA | HA | 4.16 | 7.80 | 13.0% | 4.16 | 7.80 | 13.0% | 4.16 | 7.80 | 13.0% |
| | ARIMA$_{kal}$ | 3.99 | 8.12 | 9.6% | 5.15 | 10.45 | 12.7% | 6.90 | 13.23 | 17.4% |
| | FC-LSTM | 3.44 | 6.30 | 9.6% | 3.77 | 7.23 | 10.9% | 4.37 | 8.69 | 13.2% |
| | DCRNN | 2.77 | 5.38 | 7.3% | 3.15 | 6.45 | 8.8% | 3.60 | 7.60 | 10.5% |
| | ST-GCN | 2.88 | 5.74 | 7.6% | 3.47 | 7.24 | 9.6% | 4.59 | 9.40 | 12.7% |
| | Graph WaveNet | 2.69 | 5.15 | 6.9% | 3.07 | 6.22 | 8.4% | 3.53 | 7.37 | 10.0% |
| | MRA-BGCN | **2.67** | **5.12** | **6.8%** | **3.06** | **6.17** | **8.3%** | **3.49** | **7.30** | 10.0% |
| PEMS-BAY | HA | 2.88 | 5.59 | 6.8% | 2.88 | 5.59 | 6.8% | 2.88 | 5.59 | 6.8% |
| | ARIMA$_{kal}$ | 1.62 | 3.30 | 3.5% | 2.33 | 4.76 | 5.4% | 3.38 | 6.50 | 8.3% |
| | FC-LSTM | 2.05 | 4.19 | 4.8% | 2.20 | 4.55 | 5.2% | 2.37 | 4.96 | 5.7% |
| | DCRNN | 1.38 | 2.95 | 2.9% | 1.74 | 3.97 | 3.9% | 2.07 | 4.74 | 4.9% |
| | ST-GCN | 1.36 | 2.96 | 2.9% | 1.81 | 4.27 | 4.2% | 2.49 | 5.69 | 5.8% |
| | Graph WaveNet | 1.30 | 2.74 | **2.7%** | 1.63 | 3.70 | **3.7%** | 1.95 | 4.52 | 4.6% |
| | MRA-BGCN | **1.29** | **2.72** | 2.9% | **1.61** | **3.67** | 3.8% | **1.91** | **4.46** | 4.6% |

- ARIMA$_{kal}$: Auto-Regressive Integrated Moving Average Model with Kalman filter, which is a classical time series prediction model (Hamilton, 1994).
- FC-LSTM: Recurrent neural network with fully connected LSTM hidden units (Sutskever et al., 2014).
- DCRNN: Diffusion Convolutional Recurrent Neural Network (Li et al., 2018b), which combines recurrent neural networks with diffusion convolution modeling both inflow and outflow relationships.
- ST-GCN: Spatial-Temporal Graph Convolution Network (Yu et al., 2018), which combines 1D convolution with graph convolution.
- Graph WaveNet: A convolution network architecture (Wu et al., 2019a), which introduces a self-adaptive graph to capture the hidden spatial dependency, and uses dilated convolution to capture the temporal dependency.

For all neural network based approaches, the best hyperparameters are chosen using grid search based on the performance on the validation set.

**Experimental Settings**

Recalling that the task is to learn a function $f: \mathbb{R}^{N \times P \times T'} \to \mathbb{R}^{N \times P \times T}$. In experiments, we aim at forecasting the traffic speed over one hour in the future given the traffic speed in the last hour, i.e., $T = T' = 12$.

In experiments, the number of the BGCGRU layers is set to 2, with 64 hidden units. The maximum hop $k$ of the bicomponent graph convolution is set to 3. We train our model by using Adam optimizer (Kingma and Ba 2014) to minimize the mean absolute error (MAE) for 100 epochs with the batch size as 64. The initial learning rate is 1e-2 with a decay rate of 0.6 per 10 epochs. In addition, the scheduled sampling (Bengio et al., 2015) and L2 normalization with a weight decay of 2e-4 is applied for better generalization.

Three common metrics of traffic forecasting are adopted to measure the performance of different models, including (1) Mean Absolute Error (MAE), (2) Mean Absolute Percentage Error (MAPE), and (3) Root Mean Squared Error (RMSE).

**Performance Comparison**

Table 2 presents the performances of MRA-BGCN and baseline models for 15 minutes, 30 minutes, and 1 hour ahead forecasting selected from the 12 forecasting horizons on both datasets. We observe the following phenomena:

- MRA-BGCN achieves the best performance for all forecasting horizons. It outperforms traditional traffic forecasting methods (HA, ARIMA$_{kal}$, and FC-LSTM) dramatically. MRA-BGCN also excels the vanilla GCN-based approaches (DCRNN and ST-GCN) distinctly, which perform GCN on the fixed weighted graph built according to the road network distance.
- With respect to the second-best model Graph WaveNet, we can observe that MRA-BGCN achieves small improvement on PEMS-BAY dataset, while large improvement on METR-LA dataset. From another perspective, a similar circumstance can be observed that with the growth of the forecasting horizon, the superiority of MRA-BGCN increases. Note that, the data dependency on METR-LA dataset (Los Angeles, which is known for its complicated traffic conditions) is more complicated, and long-term forecasting is inherently more uncertain than short-term forecasting. Therefore, we consider that MRA-BGCN is more capable to model complicated de-pendencies. Graph WaveNet introduces a self-adaptive graph to capture the hidden spatial dependency, which is learnt in a data-driven manner and hard to detect in complicated scenes. By contrast, we model the potential spatial dependency under the guidance of the edge interaction patterns, which can provide a better comprehension of the data and are crucial for modeling complicated dependencies.

In the following experiments, we choose to use the more complicated dataset, METR-LA.

**Effect of the Edge-Wise Graph**

To verify the effectiveness of the proposed edge-wise graph, we compare MRA-BGCN with two variants: (1) MRA-BGCN-Identity, which ignores the edge correlations and replaces the edge-wise adjacency matrix with an identity matrix. This essentially implies edges do not interact with each other and are only determined by the connected nodes; (2) MRA-BGCN-LineGraph, which replaces the edge-wise graph with the line graph, which ignores various edge interaction patterns. Table 3 shows the mean MAE, RMSE, and MAPE of 12 predictions. We can observe that, without considering the edge correlations, MRA-BGCN-Identity yields the largest testing error. Moreover, MRA-BGCN achieves the lowest testing error, which shows the effectiveness of capturing various edge interaction patterns. The intuition is that the proposed edge-wise graph considers stream connectivity and competitive relationship, and gives the model the capability of capturing complicated dependencies.

**Effect of the Multi-Range Attention Mechanism**

To further verify the effectiveness of the multi-range attention mechanism, we evaluate MRA-BGCN with the following variants using different methods for leveraging the multiple range information, including: (1) BGCN, biocomponent graph convolutional network, which ignores the multiple range information and uses representations aggregated in the given neighborhood range (i.e., only the output of layer $k$ is used); (2) MR-BGCN, multi-range bicomponent graph convolutional network, which leverages the multiple range information by concatenating representations in each layer, and considers information from each neighborhood range contributes equally. The difference between MRA-BGCN and the variants is shown in Figure 5. Table 4 shows the mean MAE, RMSE, and MAPE of 12 predictions. We can observe that BGCN, which ignores the multiple range information, achieves the worst performance, and MRA-BGCN works better than MR-BGCN. The results verify the effectiveness of the multi-range attention mechanism, which is able to leverage multiple range information and distinguish the importance of different neighborhood ranges.

Table 3: The Performance Comparison of MRA-BGCN and MRA-BGCN without Edge-wise Graph

| models | MAE | RMSE | MAPE |
| --- | --- | --- | --- |
| MRA-BGCN-Identity | 3.13 | 6.33 | 8.8% |
| MRA-BGCN-LineGraph | 3.07 | 6.10 | 8.6% |
| MRA-BGCN | **3.05** | **6.04** | **8.3%** |

Table 4: The Performance Comparison of MRA-BGCN and MRA-BGCN without the Multi-range Attention Mechanism

| models | MAE | RMSE | MAPE |
| --- | --- | --- | --- |
| BGCN | 3.08 | 6.19 | 8.9% |
| MR-BGCN | 3.07 | 6.13 | 8.8% |
| MRA-BGCN | **3.05** | **6.04** | **8.3%** |

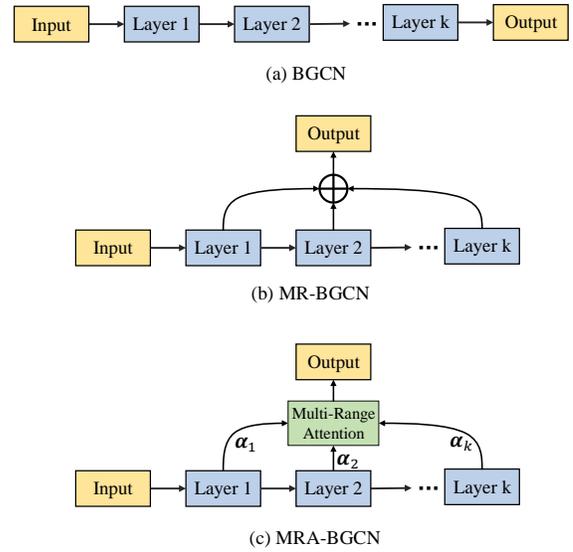

(a) BGCN

(b) MR-BGCN

(c) MRA-BGCN

Figure 5: The Illustration of MRA-BGCN and the Variants

## Conclusions and Future Work

We propose the *Multi-Range Attentive Bicomponent Graph Convolutional Network* for traffic forecasting. Specifically, the bicomponent graph convolution is proposed to explicitly model the correlations of both nodes and edges. An edge-wise graph construction approach is proposed to encode stream connectivity and competitive relationship. The multi-range attention mechanism is proposed to efficiently leverage multiple range information and generate integrated representations. On two traffic datasets, our model achieves the state-of-the-art performance. For future work, we will investigate the following two aspects (1) applying the proposed model to other spatial-temporal forecasting tasks; (2) extending our approach to model more complex spatial-temporal dependencies considering more factors, e.g., traffic accidents and surrounding points of interest.

# Acknowledgement

This work is supported by the National Key Research and Development Program of China (No. 2018YFB0505000).